\documentclass[a4paper, 
10pt
]{article}

\usepackage[a4paper,left=3cm,right=3cm,top=2.5cm,bottom=2.5cm]{geometry}
\usepackage{amsmath}
\usepackage{multirow}
\usepackage[table]{xcolor}
\usepackage{amsopn}

\newcommand{\A}[1]{\color{black}{ #1}\color{black}}
\usepackage{subcaption}
\usepackage{pgfplots}
\usepackage{pgfplotstable}
\usepackage{mathtools}
\usepackage{multicol}
\usepackage{comment}
\usepackage{booktabs}
\usepackage{amssymb}
\usepackage{url}
\usepackage{bm}
\usepackage[a4paper,left=3cm,right=3cm,top=2.5cm,bottom=2.5cm]{geometry}
\usepackage{amsmath}
\usepackage{multirow}
\usepackage{float}
\usepackage{tabularx}
\usepackage{enumerate}
\usepackage{array}
\usepackage{xspace}
\usepackage{tikz}
\usepackage[export]{adjustbox}
\usepackage{tikzsymbols}
\usepackage{authblk}
\usepackage{mathrsfs}

\numberwithin{equation}{section}
\usepgfplotslibrary{groupplots}
\usepgfplotslibrary{fillbetween}
\pgfplotsset{compat=1.5}
\usetikzlibrary{calc,trees,positioning,arrows,chains,shapes.geometric,%
    decorations.pathreplacing,decorations.pathmorphing,shapes,%
    matrix,shapes.symbols, decorations.markings, patterns,fit,quotes,%
    arrows.meta, spy}

\usepackage{xcolor}
\definecolor{Gray}{gray}{0.9}

\usepackage{hyperref}
\usepackage{siunitx}
\usepackage{cleveref}
\usepackage{algorithm}
\usepackage{algpseudocode}
\usepackage[bottom]{footmisc}

\graphicspath{{figures/}}

\newcommand{\mupar}{\ensuremath{\boldsymbol{\mu}}}

\newcommand{\y}{\ensuremath{\mathbf{y}}}

\newcommand{\norm}[1]{\lVert #1 \rVert}

\usepackage[english]{babel}
\usepackage{amsthm}

\newtheorem*{remark}{Remark}

\begin{document}

\title{An extended physics informed neural network for preliminary analysis of parametric optimal control problems}

\author[]{Nicola~Demo\footnote{nicola.demo@sissa.it}}
\author[]{Maria Strazzullo\footnote{maria.strazzullo@sissa.it}}
\author[]{Gianluigi~Rozza\footnote{gianluigi.rozza@sissa.it}}

\affil{Mathematics Area, mathLab, SISSA, via Bonomea 265, I-34136
  Trieste, Italy}

\maketitle

\begin{abstract}
In this work we propose an extension of physics informed supervised learning strategies to parametric partial differential equations. Indeed, even if the latter are indisputably useful in many applications, they can be computationally expensive most of all in a real-time and many-query setting. Thus, our main goal is to provide a physics informed learning paradigm to simulate parametrized phenomena in a small amount of time. The physics information will be exploited in many ways, in the loss function (standard physics informed neural networks), as an augmented input (extra feature employment) and as a guideline to build an effective structure for the neural network (physics informed architecture). These three aspects, combined together, will lead to a faster training phase and to a more accurate parametric prediction. The methodology has been tested for several equations and also in an optimal control framework.
\end{abstract}


\section{Introduction}
\label{sec:intro}
Machine Learning represents a growing impact research topic, widespread in several fields of applications, see for example \cite{ding2001multi,krizhevsky2012imagenet,lin2018all, wang2004modeling}. This massive improvement and employment of such a technique is related to the growing disposal of available data and computing resources. Yet, the analysis of a complex system needs a huge quantity of data in order to be reliable and meaningful. However, collecting data to train can be unbearable for many reasons. Indeed, the costs of the collection can be expensive and data usually feature scattered information. This lack of knowledge might give non-robust results that can be inconsistent with respect to the known behaviour of a physical phenomenon. In this context, in the last few years, the Physics Informed Neural Networks (PINNs) have been conceived and developed in order to tackle this issue. The main idea behind PINNs is to add some physical information to the Neural Network (NN) in order to reach more reliable and complete predictions. The term physics usually translates to a mathematical model, such as, in our case, partial differential equations (PDEs). For an overview on the topic, the interested reader might refer to \cite{raissi2019physics}. The promising results of this seminal paper, paved the way to many applications and extensions, see, e.g., this far-from-exhaustive list \cite{cai2021physics, jagtap2020conservative, jin2021nsfnets, kharazmi2021hp,mahmoudabadbozchelou2021data, meng2020ppinn, pang2020npinns,raissi2020hidden, yang2021b}, where PINNs and its variations have been applied to many fields.
This work focuses on the need of reliably predict physical phenomena in a parametric setting for complex systems. Motivated and inspired by~\cite{Lu2021} \A{and the latest works~\cite{A2,A1}}, we want to find a PINN-based strategy to deal with equations where a parameter might change the physical features or the properties of the model itself. In many industrial and scientific contexts, there is a need of solving parametric PDEs (PDE(\mupar)s) for many values of a parameter $\mupar$. Indeed, we have in mind the need for fast simulations in \emph{real-time} and \emph{many-query} applications, where $\mupar$-dependent solutions of a PDE($\mupar$) are required in a small amount of time. Thus, we propose to adapt the PINN paradigm to a parametric setting. As already mentioned, the PINNs are a valuable tool to predict a natural phenomenon according to a mathematical model, not relying on collected data information. However, the model equations, in several fields, are not sufficient to completely describe and understand a natural phenomenon. Is there a way to change the system at hand in order to achieve a \emph{desired configuration}, which can represent expected behaviour extrapolated from scattered \emph{in situ} data collection or a convenient profile to reach? Towards this goal, PDE($\mupar$)s constrained optimal control problems (OCP($\mupar$)s) can be employed in order to steer the problem solution towards a given configuration. Furthermore, we are interested in those applications where both the equation and the desired profile are parameter dependent. In this parametric setting, OCP$(\mupar)$s have been successfully employed in fluid dynamics, see e.g.\ \cite{de2007optimal, dede2007optimal, negri2015reduced, povsta2007optimal, FedeMaria}, and in many scientific applications, such as biomedical \cite{ballarin2017numerical, lassila2013reduced, ZakiaMaria, Zakia} and environmental ones \cite{CARERE2021261,quarteroni2005numerical, quarteroni2007reduced,Strazzullo1,Strazzullo3}. The main goal of the previous works was the application of model order reduction \cite{hesthaven2015certified,prud2002reliable} to provide a low dimensional function space to solve the parametric instances in a faster way with respect to standard mesh-based approximation techniques such as Finite Elements, Finite Volumes, Spectral Methods, see for example \cite{quarteroni2008numerical}. In this work, we want to exploit PINN as a complete machine learning-based way to predict OCP($\mupar$)s solutions. It is clear that more complex problems are related to a possibly time consuming training phase. To accelerate this procedure, some ruses have been conceived, it is the case of activation functions \cite{jagtap2020locally,jagtap2020adaptive} or Prior Dictionary PINNs \cite{peng2020accelerating}. The strategy we propose is based on exploiting additional known information, the \emph{ures}, in the input of the NN: this will allow to converge faster to a loss minimum not paying in the net prediction accuracy. Furthermore, we propose a different interpretation of the PINN paradigm, where the physics information is not only used to force the system to reach the expected model behaviour, but also to change the structure of the NN one is dealing with.
\A{The main novelty of this work, to the best of our knowledge, is the numerical investigation of tailored parametric PINN strategy for OCP($\mupar$)s. The application of PINN to optimal control problems has been recently investigated in \cite{MowlaviNabi}. However, differently from \cite{MowlaviNabi}, we focus on the benefits of exploiting physics informed structures and extra features as a general strategy to deal with system of multiple equations, most of all in the parametric setting. Moreover, we combine the physics informed structures with extra features as an augmented input to accelerate the training procedure. We stress that parametric PDEs have been successfully tackled in \cite{A2, A1, Lu2021,wang2021learning}. However, in this contribution, we propose a different approach that totally complies with the original PINNs formulation of \cite{raissi2019physics} with very encouraging results. This aspect is very important in terms of coding and applicability since the proposed approaches can be easily implemented for any PINN-code.}
\\
\A{The work is outlined as follows: in Section \ref{sec:methodology}, we will introduce the PINN structure in a parametric setting for one dimensional output and multi-dimensional outputs. In the same Section we will also present the main idea behind the employment of the extra features to accelerate the training phase. 
Section \ref{sec:results_ocp} represents a first step of the employment of PINNs for multi-variable output in the context of OCP($\mupar$)s, for Poisson and Stokes equations. Finally, conclusions follow in Section \ref{sec:conclusions}. Moreover, we present some additional results in \ref{sec:results_laplace_burgers}, where we validate the proposed methodology on forward one-dimensional problems such as Poisson equation \cite{Atangana2013} and Burgers equation \cite{raissi2019physics}. The numerical tests have been tackled both in a parametric and non-parametric formulation. }


\section{Methodology}
\label{sec:methodology}
This Section focuses on the various methodologies we relied on to tackle several PDE-based problems. First of all, we will describe PINNs-based methods in Section \ref{sec:PINN}. As already mentioned in Section \ref{sec:intro}, one of the main novelty of the work is their extension to parametrized OCP($\mupar$)s. Section \ref{sec:features} concerns the employment of \emph{extra-features} that will help the convergence of the neural network, reducing the computational time needed to train the model.
Finally, we will also present the how to exploit the physics information to change the architecture of the used PINN to have more accurate results in Section \ref{sec:PINN_multiple_eq}. This will be helpful when dealing with systems of multiple equations. This technique, already successfully employed for multi--fidelity problems~ \cite{baydin2018automatic,guo2021multi,motamed2020multi}, allowed us to reach better results with respect to standard PINN approach.
\subsection{PINNs}
\label{sec:PINN}
Let us assume to be provided by a PDE($\mupar$) endowed with its boundary conditions of the form $G: \mathbb Y \rightarrow \mathbb Q^*$ that reads \begin{equation}
\label{eq:G_mu}
G(w(\mathbf x, \mupar)) = f(\mathbf x, \mupar),
\end{equation}
where $w \doteq w(\mathbf x, \mupar) \in \mathbb Y$ is the unknown physical quantity we are taking into consideration in a domain $\Omega \subset \mathbb R^d$, while $f(\mupar) \in \mathbb Q^*$ is an external forcing term, with $\mathbb Y$ and $\mathbb Q$ two suited Hilbert spaces. The system changes with respect to a parameter $\mupar \in
\mathcal D \subset \mathbb R^D$, of dimension $D \geq 1$. The parameter $\mupar$ represents physical and/or geometrical features of the problem at hand. In this work, we will focus on physical parametrization only. With this notation, we are referring to a broad class of PDE($\mupar$)s: from time-dependent nonlinear problems, to linear steady ones and to coupled systems (as OCP(\mupar)s). Indeed, when dealing with time-dependent problems, we are implicitly considering a time evolution of $w$ in the time interval $[0,T]$ in the expression \eqref{eq:G_mu}. Thus, depending on the context, the variable $\mathbf x$ can be interpret as $\mathbf x \doteq x$ for steady problems and as $\mathbf x \doteq (x, t)$ in the time-dependent case, where $x \in \mathbb R^d$ represents the spatial coordinate vector while $t \in [0,T]$ is the time variable. Along this contribution we will also consider PDEs that do not depend on a parameter, in that case   $G: \mathbb Y \rightarrow \mathbb Q^*$ reads 
\begin{equation}
\label{eq:G}
G(w(\mathbf x)) = f(\mathbf x).
\end{equation}
The same argument for time-dependent problems applies to the non-parametric context of \eqref{eq:G}.
We aim at finding data-driven solutions to \eqref{eq:G_mu} or \eqref{eq:G}. For the sake of brevity we will describe the methodology only in the parametric case. Indeed, this more general formulation, will easily adapt to the context of non-parametric PDEs. Building on the strategy firstly conceived in \cite{raissi2019physics}, we propose a PINN able to capture the value of the solution $w$ for several parametric instances. To this end, we define the residual of $\eqref{eq:G_mu}$ as: 
\begin{equation}
\label{eq:res_mu}
r(w(\mathbf x, \mupar)) \doteq G(w(\mathbf x, \mupar)) - f(\mathbf x, \mupar).
\end{equation}
Now let us imagine to have been provided of some boundary values, say $N_{b}$ points $\{\mathbf x_i^b\}_{i=1}^{N_{b}}$, of $N_p$ internal points $\{\mathbf x_i^p\}_{i=1}^{N_{p}}$ and of $N_{\mupar}$ points in the parameter space, i.e.\ $\{\mupar_i\}_{i=1}^{N_{\mupar}}$. We want to exploit the data information given by the residual in a NN $\hat{w}$ that takes as input some domain coordinates and a parametric instance and gives back a prediction of the physical phenomenon. In this sense, the network will be \emph{physics informed} and will exploit the residual value in order to converge towards a physical meaningful solution $w$. The differential structure of the residual can be derived by applying automatic differentiation based on the chain rule for
differentiating compositions of functions \cite{baydin2018automatic, raissi2019physics}. In order to embed the physical behaviour in the NN at hand, we define the following mean squared error loss:
\begin{equation}
\label{eq:global_loss}
    MSE^{\mupar}  \doteq \frac{1}{N_{\mupar}}\sum_{i=1}^{N_{\mupar}} ( MSE_b^{\mupar_i} + MSE_p^{\mupar_i}),
\end{equation}
where 
\begin{itemize}
    \item[$\circ$] given the boundary values of the solution $w_i^b$ for $i=1, \dots, N_b$ at the collocation points $\{\mathbf x_i^b\}_{i=1}^{N_{b}}$ the \emph{boundary loss} is
    \begin{equation}
        MSE_b^{\mupar_i} \doteq \frac{1}{N_b} \sum_{k=1}^{N_b} | \hat{w}(\mathbf x_k^b, \mupar_i) - w(\mupar_i)_k^b|^2;
    \end{equation}
    \item[$\circ$] while the \emph{residual loss} is:
    \begin{equation}
        MSE_{p}^{\mupar_i} \doteq \frac{1}{N_p}\sum_{k=1}^{N_p}|r(\hat{w}(\mathbf x_k, \mupar_i ))|^2.
    \end{equation}
\end{itemize}
In this context, $(\mathbf x_k^b, w(\mupar_i)_k^b)$ for $k=1, \dots, N_b$ are the boundary training data that might be $\mupar-$dependent, while $\{\mathbf x_i^p\}_{i=1}^{N_{p}}$ are the collocation point for the $\mupar-$dependent residual. Namely:
\begin{itemize}
    \item[$\circ$] the first set of inputs enforces the boundary conditions in space and, possibly, in time,
    \item[$\circ$] the residual evaluated in the internal points is informative about the structure of the physical model one is interested in.
\end{itemize}
To the best of our knowledge, it is the first time that the parameters are considered as inputs of the model, together with the points of the domain in space and time. Indeed, in seminal paper about PINN, see e.g. \cite{raissi2019physics} the parameters of the problem were fixed or the PINN was used in parameter identification, also in an inverse problem Bayesian setting \cite{yang2021b}. Instead, in this contribution, inspired by \cite{Lu2021}, we want to take a first step in the employment of PINNs in many-query contexts, where several parameter evaluations are required to deeper analyse the physical phenomenon at hand. The main advantage of this parametric approach relies in the versatility of the considered model, where many configurations can be learnt from a finite parametric training data $\{\mupar_i\}_{i=1}^{N_{\mupar}}$. The PINN can respond, in a total non-intrusive way, to the need of reliable solutions of PDE(\mupar)s for several values of $\mupar$ in \emph{real-time} and \emph{many-query} contexts, where many evaluations of the equation \ref{eq:G_mu} are required in a small amount of time. Indeed, paying a reasonable amount of time training the net, it is possible to fast predict faithfully a new parametric instance.
\begin{remark}[The non-parametric case]
\label{remark:non_mu}
It is clear that, for the non parametric case, we are dealing with the structure first presented in \cite{raissi2019physics}. For the sake of clarity, we report the structure we used in this simplified setting. First of all, the residual is defined as $MSE \doteq MSE_b + MSE_p$, where 
\begin{equation}
\label{eq:res_mse}
 MSE_b \doteq \frac{1}{N_b} \sum_{k=1}^{N_b} | \hat{w}(\mathbf x_k^b) - w_k^b|^2 \quad \text{and} \quad MSE_p \doteq \frac{1}{N_p}\sum_{k=1}^{N_p}|r(\hat{w}(\mathbf x_k))|^2,
\end{equation}
where, naturally, from \eqref{eq:G}, the non-paramtric residual is given by
\begin{equation}
\label{eq:res}
r(w(\mathbf x)) \doteq G(w(\mathbf x)) - f(\mathbf x).
\end{equation}
Namely, in this case, it is sufficient to be provided of $N_{b}$  boundary points $\{\mathbf x_i^b\}_{i=1}^{N_{b}}$ with the related PDE boundary values $w_k^b$ together with $N_p$ internal points to evaluate \eqref{eq:res}, since the NN $\hat{w}$ takes input only the domain coordinates $\mathbf x$.

\end{remark}
\subsection{Extra Features}
\label{sec:features}
By definition, NNs are a composition of nonlinear functions.
Intuitively this implies that the greater the number of functions involved within the composition is, the more complex the output of the network will be. Previous works~\cite{jagtap2020adaptive,jin2021nsfnets} have shown that increasing the depth of the network allows to learn PDEs with a high nonlinearity, indeed. On the other hand, increasing the number of hidden layers --- usually equal to the number of activation functions --- makes the training phase even more expensive from the computational viewpoint and may introduce other numerical problems, e.g. vanishing gradient.

One possible way to contain the dimension of the model and preserving at the same time the output complexity can be the extension of the features we use as input of the network. In a typical PINN framework, the features are the dimensions of the input domain, i.e. the spatial and temporal coordinates, and, in case of parametric problems, also the parameters. For example in an unsteady parametric problem the input is $\mathbf x = \begin{bmatrix}x_1 & \dots & x_d & \mu_1 & \dots & \mu_D &t\end{bmatrix}$.
We can exploit our a priori knowledge about the physical equations to extend the information we use as input, evaluating one (or more) input variable using one (or more) kernel function\footnote{which should be differentiable.}. We define these functions as $\{k_i(\mathbf x)\}_{i=1}^{N_f}$ and we will refer to them as \emph{extra features} from here on. The new input is then 
\begin{equation}
\mathbf x_\text{extra} = 
\begin{bmatrix}
  \mathbf{x} & 
  k_1(\mathbf{x}) & \dots & k_{N_f}(\mathbf{x})
  \end{bmatrix}
\end{equation}
of dimension $d+D+1+{N_f}$.
Of course, also functions defined into subspaces of the input space are valid as well. For a concrete example, let us imagine to dealing with a generic problem that shows a quadratic dependency by the first spatial dimension and an exponential behaviour in time: a reasonable input should be $\mathbf x_\text{extra} = \begin{bmatrix}x_1 & \dots & x_d & t & x_1^2 & e^t\end{bmatrix}$. Selecting a proper set of extra features, the model will learn an easier correlation between input and output, resulting in a faster training and a smaller architecture. On the practical side, the extra features are problem dependent and they have to be tuned to avoid loss of performances. We present here only few guidelines to select good features, postponing the deeper discussion regarding their effectiveness to Section \ref{sec:results}. In case of problems with external forces, the analytical function representing such force could be a valid option for an extra feature. In this way, the NN uses the new feature as a sort of initial guess of its learning procedure. 
Similarly, simple functions that satisfy the boundary (or initial) conditions could lead the training procedure to been improved.
We remark that we can moreover adding learnable parameters within the extra features, allowing for adaptive extra features that, as we will discuss in Section~\ref{sec:results}, improve the training step both in terms of accuracy and computational needed.
\subsection{PINNs for multiple equations problems}
\label{sec:PINN_multiple_eq}
In this Section we are going to extend the idea of using a NN to approximate the physical behaviour given by system of multiple equations. Namely, we are facing a problem of the form
\begin{equation}
\label{eq:G_mu_coupled}
\mathcal G(w(\mathbf x, \mupar)) = \mathcal F(\mathbf x, \mupar),
\end{equation}
where the solution is made of $n$ variables $(w_1(\mathbf x, \mupar), \dots, w_n(\mathbf x, \mupar))$, while the left and the right hand side are defined as:
\begin{equation}
    \mathcal G(w(\mathbf x, \mupar)) \doteq
    \begin{bmatrix}
    G_1(w(\mathbf x, \mupar))\\
    \quad \quad \;\vdots \\
    G_n(w(\mathbf x, \mupar))\\
    \end{bmatrix}
    \quad \text{and} \quad 
    \mathcal F(\mathbf x, \mupar) \doteq
    \begin{bmatrix}
    f_1(\mathbf x, \mupar)\\
    \quad \quad \;\vdots \\
    f_n(\mathbf x, \mupar)\\
    \end{bmatrix}.
\end{equation}
It is clear that, in order to solve the problem \eqref{eq:G_mu_coupled} in a PINN fashion, a new definition of the mean squared error loss is needed, to take into account all the involved equations and boundary conditions. First of all we define the residual of the $j-$th equation as
\begin{equation}
\label{eq:res_j_mu}
r_j(w(\mathbf x, \mupar)) \doteq G_j(w(\mathbf x, \mupar)) - f_j((\mathbf x, \mupar)).
\end{equation}
We recall that problem \eqref{eq:G_mu_coupled} is related to a set of boundary conditions. Thus, let $J = \{1, \dots, m\}$ be the indices of the number of boundary conditions applied while let us define $I = \{1, \dots, n\}$. Now, given $\mupar \in \mathcal D$ and the related sets of boundary points $\{\mathbf {x}_{i}^b\}_{i=1}^{N_b^l}$ and boundary conditions $\{{w(\mupar)_i^b}\}_{i=1}^{N_b^l}$ for $l \in J$ we can define:
\begin{itemize}
    \item[$\circ$] the new boundary loss as
    \begin{equation}
    \label{eq:MSE_b}
        MSE_b^{\mupar_i} \doteq \sum_{l \in J}
        \frac{1}{N_b^l} \sum_{k=1}^{N_b^l} | \hat{w}(\mathbf x_k^b, \mupar_i) - w(\mupar_i)_k^b|^2;
    \end{equation}
    \item[$\circ$] the new residual loss as
        \begin{equation}
        \label{eq:MSE_p}
        MSE_{p}^{\mupar_i} \doteq \sum_{j \in I} \frac{1}{N_p}\sum_{k=1}^{N_p}|r_j(\hat{w}(\mathbf x_k, \mupar_i ))|^2,
    \end{equation}
\end{itemize}
where $\hat{w}$ is a NN capable to approximate the global solution of the system of equations considered.
At the end, the global loss can be exactly defined as we already did in \eqref{eq:global_loss}. The problem can be solved through the PINN approach already presented in Section~\ref{sec:PINN} and the extra features arguments also applies and easily adapts in this context. Furthermore, this structure can be easily specified to the non-parametric case as presented in Remark \ref{remark:non_mu}.
In the next Section we are going to question ourselves on the possibility of building a \emph{physics informed architecture} (PI-Arch) to better deal with problems with more than one variable. In this framework, the model does not only influence the loss function, but also the structure of the NN itself. 

\subsection{The PI-Arch strategy for multiple equations problems}
This Section focuses on the idea of exploiting the PINN paradigm in a more complete way. Indeed, once provided of some physical information about the problem at hand, it can be exploited not only in the loss, but also in the structure of the employed NN. We are going to extend the idea of using a NN to approximate the physical behaviour given by system of multiple equations. The idea builds on multi--fidelity approaches as used in~ \cite{baydin2018automatic,guo2021multi,motamed2020multi} and it was guided by the numerical tests we performed on OCP($\mupar$)s. The comment on the numerical results are postponed in Section \ref{sec:ocp}. We now focus on a general description of the adopted strategy. We recall that we are working in the framework of multiple variables described by system \eqref{eq:G_mu_coupled}. Namely we are dealing with $n$ equations and $m$ boundary conditions. First of all, the physics is taken into account in the loss definition, indeed the global loss is defined as in \eqref{eq:global_loss}, where $MSE^{\mupar}_p$ and $MSE^{\mupar}_b$ are exactly defined as in \eqref{eq:MSE_p} and \eqref{eq:MSE_b}. Nonetheless, we stress that the same arguments of Remark \ref{remark:non_mu} apply also in this more complicated context with analogous definitions.\\
The idea is to change the architecture of the NN $\hat w$ exploiting the structure of the model behind the phenomenon we are studying. Namely, we would like to adapt the structure in order to be more accurate in the prediction of the outputs. Let us suppose to have built a tailored (combination of) NN (NNs) to approximate the PDE($\mupar$) solution.
Let us consider $H_1 \subset I$. Namely, a first NN takes the inputs and predicts a part of the variables $\{\hat{w}_{h_1}(\mathbf x, \mathbf \mu)\}_{{h_1} \in H_1}$. This latter set of predicted solution will be called the \emph{one-level output} (1-out). An improper subset of the one-level output together with the initial input (or a part of it) will be the input for a new NN which will predict the \emph{two-level output} (2-out) $\{\hat{w}_{h_2}(\mathbf x, \mathbf \mu)\}_{h_2 \in H_2}$, where $H_2 \subset I \setminus H_1$. These output variables combined with the one-level output (or a part of it) and the initial input (or part of it) will lead to the \emph{three-level output} (3-out) for the set of indices $H_3\subset I\setminus \{H_1 \cup H_2\}$, and so on. The process is repeated until the $k-$level, where the final output will be $\{\hat{w}_{h_k}(\mathbf x, \mathbf \mu)\}_{{h_k} \in H_k}$, with
$$
\bigcup_{k} H_k = I.
$$
Now let us image that from the equations, directly or indirectly, we can derive a sort of relation, say $\Xi$, between the outputs. Namely, we are able to connect 1-out to the 2-out through
\begin{equation}
\label{eq:xi_relation}
    \Xi(\text{1-out}) \approx \text{2-out}. 
\end{equation}
Thus, this information pave the way to two different strategies:
\begin{itemize}
    \item[$\circ$] if confident enough, replace the NN connecting the 1-out to the 2-out with $\Xi(\text{1-out})$;
    \item[$\circ$] build a NN capable to approximate $\Xi(\text{1-out})$.
\end{itemize}
A schematic representation of the proposed architecture is given in Figure~\ref{fig:PINN_cascade}.
 Such approach aims to divide the NN in hand into a combination of different (typically smaller) NNs, thus isolating the relations between the outputs of the model --- that we know a priori from the physics equations --- that otherwise have to be learned during the training. In this way, we exploit the physical information not only for the loss definition, but also in the structure of the model. For this reason, we refer to this strategy in the rest of the manuscript as \emph{physics informed architecture} (PI-Arch). 

\def\layersep{1.2cm}
\begin{figure}[H]
\begin{center}
\begin{tikzpicture}[shorten >=1pt,->,draw=black!50, node distance=\layersep]
    \tikzstyle{every pin edge}=[<-,shorten <=1pt]
    \tikzstyle{neuron}=[circle,fill=black!25,minimum size=10pt,inner sep=0pt]
    \tikzstyle{input neuron}=[neuron, fill=cyan!30];
    \tikzstyle{output neuron}=[neuron, fill=magenta!40];
    \tikzstyle{hidden neuron}=[neuron, fill=teal!50];
    \tikzstyle{annot} = [text width=4em, text centered]

    \foreach \name / \y in {1,...,3}
        \node[input neuron, pin=left:{\it \footnotesize input \#\y}] (I1-\name) at (0,-\y) {};

    \foreach \name / \y in {1,...,4}
        \path[yshift=0.5cm]
            node[hidden neuron] (H1-\name) at (\layersep,-\y cm) {};

    \node[output neuron,pin={[pin edge={->}, align=left]right:{\it \footnotesize {1-out}}}, right=2cm of I1-2] (O1) {};

    \node[input neuron, pin=left:{~}, minimum size=0pt,right=2.2cm of O1] (out) {};

    \foreach \source in {1,...,3}
        \foreach \dest in {1,...,4}
            \path (I1-\source) edge (H1-\dest);

    \foreach \source in {1,...,4}
        \path (H1-\source) edge (O1);
        
    \node[input neuron, right=4cm of O1, pin=left:{\it \footnotesize $\Xi(\text{1-out})$}] (I-1) {};
    \node[input neuron, above=0.7cm of I-1, pin=left:{\it \footnotesize input \#4}] (I-2) {};
    \node[input neuron, below=0.7cm of I-1, pin=left:{\it \footnotesize input \#5}] (I-3) {};

    \foreach \name / \y in {1,...,4}
        \path[yshift=0.5cm]
            node[hidden neuron] (H-\name) at (190+\layersep,-\y cm) {};

    \foreach \source in {1,...,3}
        \foreach \dest in {1,...,4}
            \path (I-\source) edge (H-\dest);
    \node[output neuron,pin={[pin edge={->}, align=left]right:{\it \footnotesize $2$--out}}, right=2cm of I-1] (O) {};

    \foreach \dest in {1,...,4}
        \path (I-1) edge (H-\dest);

    \foreach \source in {1,...,4}
        \path (H-\source) edge (O);

    \node[annot,above of=H-1, node distance=1cm] (hl1) {$2^{nd}$ NN};
    \node[annot,above of=H1-1, node distance=1cm] (hlk) {$1^{st}$ NN};

\end{tikzpicture}
\end{center}
\caption{PI-Arch example.}
\label{fig:PINN_cascade}
\end{figure}
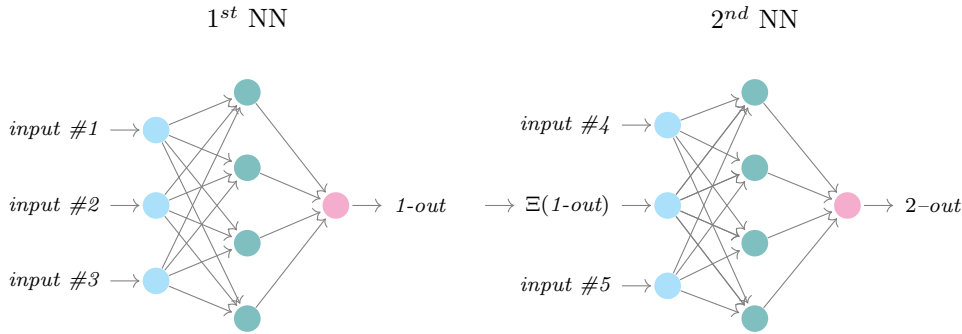

\section{Some First Numerical Results}
\label{sec:results}

\section{PINNs for Optimal Control Problems}
\label{sec:results_ocp}
In this Section we are going to test our methodology in a more complex setting, where a coupled system of equations is solved. We will deal with OCP($\mupar$)s: this mathematical tool is used to change the classical behaviour of a physical variable to reach a given desired configuration. Optimal Control framework is thus an input-output system that, given an observable, tries to drive the solution towards this data thanks to an external variable called \emph{control} \cite{bochev2009least, gunzburger2003perspectives, hinze2008optimization, lions1971, troltzsch2010optimal}. OCP($\mupar$)s have been employed in many applications in several scientific and industrial fields, the interested reader might refer to \cite{leugering2014trends} for an overview. In the next Section we will introduce the problem setting and we will show some numerical results based on PI-Arch application.

\subsection{Problem Formulation}
\label{sec:ocp}
Here, we here provide the continuous formulation for general OCP($\mupar$)s. Indeed, the following setting is suited to nonlinear problem as well as time dependent ones. However, we will focus on steady problems: the choice is guided by the numerical results we will discuss in the Section. After introducing the problem at hand, we rely on the Lagrangian formalism \cite{gunzburger2003perspectives, hinze2008optimization,ito2008lagrange} to solve the constrained minimization problem.
Let $\Omega \subset \mathbb R^d$, with $d=2,3$ be the spatial domain where the physical phenomena we are interested in is occurring. We assume to be provided by \emph{state equation} $G: \mathbb Y \rightarrow \mathbb Q^*$ of the form \begin{equation}
\label{eq:state}
G(y(\mathbf x, \mupar)) = f(\mathbf x, \mupar).
\end{equation}
We call $y(\mathbf x, \mupar)$ the \emph{state variable}, i.e.\ the physical variable we want to steer towards a desired profile $y_\text{d} \doteq y_\text{d}(\mupar) \in {\mathbb Y}_{\text{obs}} \supseteq \mathbb Y$. Also in this case, $f(\mathbf x, \mupar) \in \mathbb Q ^*$ is an external forcing term. We now denote with the notation $\mathcal L(\cdot, \cdot)$ the space of the continuous linear functions between two function spaces. As already specified, the aim of the problem, i.e.\ to steer the solution behaviour towards a desired profile, is reached through a \emph{control variable} $u(\mathbf x, \mupar) \in \mathbb U$. Here, $\mathbb U$ is another Hilbert space. The control variable acts in $\Omega_{u} \subset \Omega$, the \emph{control domain}. In this work, we will assume $\Omega_u = \Omega$, namely, we will test the proposed methodology for \emph{distributed} OCP($\mupar$)s. In order to change the classical solution behaviour, we need to define a \emph{controlled system}, i.e.\
$\mathcal E: \mathbb Y \times \mathbb U  \rightarrow \mathbb Q^*$ such that:
\begin{equation}
\label{eq:control_eq}
\mathcal E(y(\mathbf x, \mupar),u(\mathbf x, \mupar)) \doteq\; G(y(\mathbf x, \mupar)) - C(u(\mathbf x, \mupar)) - f(\mathbf x, \mupar)= 0.
\end{equation}
The operator $C \in \mathcal L(\mathbb U, \mathbb Q ^*)$ changes the original system to reach the desired variable $y_\text{d}$. Theoretically, this goal is represented by the solution of the following constrained minimization problem: given a $\mupar \in \mathcal D$ and an observation $y_{\text{d}} \in \mathbb Y_{\text{obs}},$ find the pair $(y(\mathbf x, \mupar),u(\mathbf x, \mupar)) \in \mathbb Y \times \mathbb U$ which solves
\begin{equation}
\label{eq:min_problem}
\min_{y(\mathbf x, \mupar) \in {\mathbb Y}, u(\mathbf x, \mupar) \in {\mathbb U}}J(y(\mathbf x, \mupar),u(\mathbf x, \mupar); y_\text{d}(\mathbf x, \mupar)) \text{ such that } \mathcal E(y(\mathbf x, \mupar),u(\mathbf x, \mupar)) = 0,
\end{equation}
where $J: \mathbb Y \times \mathbb U \times {\mathbb Y}_{\text{obs}} \rightarrow \mathbb R$ a \emph{cost functional} of the form:
\begin{equation}
J(y(\mathbf x, \mupar),u(\mathbf x, \mupar); y_\text{d}) \doteq \frac{1}{2} \norm{y(\mathbf x, \mupar) - y_\text{d}(\mathbf x, \mupar)}_\mathbb {Y_{\text{obs}}}^2 + \frac{\alpha(\mupar)}{2} \norm{u(\mathbf x, \mupar)}_{\mathbb U}^2,
\end{equation}
where $\alpha(\mupar) \in (0, 1]$ is a \emph{penalization term}. The reader interested in the well-posedness of the model may refer to \cite{hinze2008optimization} for example. 
The optimal control problem can be solved through a Lagrangian argument  \cite{hinze2008optimization,troltzsch2010optimal}, defining an arbitrary \emph{adjoint variable} $z(\mathbf x, \mupar)$ in $\mathbb Y$. The use of this variable will allow us to solve the problem \eqref{eq:min_problem} in an unconstrained minimization fashion. In order to recover the notation we employed in the previous Sections, we define the \emph{global variable} $w(\mathbf x, \mupar) \doteq (y(\mathbf x, \mupar),u(\mathbf x, \mupar),z(\mathbf x, \mupar))$. Then, we can define the following \emph{Lagrangian functional}
\begin{align*}
\label{lagrangian_functional}
\mathscr L(w(\mathbf x, \mupar); y_\text{d}(\mathbf x, \mupar)) & \doteq J(y(\mathbf x, \mupar), u(\mathbf x, \mupar); y_\text{d}(\mathbf x, \mupar))  \\
& \quad \quad \quad \quad \quad + \langle z(\mathbf x, \mupar), \mathcal E(y(\mathbf x, \mupar),u(\mathbf x, \mupar)) \rangle_{\mathbb Q \mathbb Q^*},
\end{align*}
where $\langle \cdot, \cdot \rangle_{\mathbb Q\mathbb Q^*}$ is the duality pairing of the spaces $\mathbb Q$ and $\mathbb Q^*$ and we are assuming $\mathbb Y \subset \mathbb Q$. In this setting, it is well-known in literature \cite{bochev2009least, gunzburger2003perspectives, hinze2008optimization, lions1971, troltzsch2010optimal} that the minimization problem \eqref{eq:min_problem} is equivalent to the solution $(y(\mathbf x, \mupar), u(\mathbf x, \mupar), z(\mathbf x, \mupar)) \in \mathbb Y \times \mathbb U \times \mathbb Y$ of the following \emph{optimality system}:
\begin{equation}
\label{eq:KKT}
\begin{cases}
D_{y}\mathscr L(w(\mathbf x, \mupar); y_\text{d}(\mathbf x, \mupar)) [\omega] = 0 & \forall \omega \in \mathbb Y,\\
D_u\mathscr L(w(\mathbf x, \mupar); y_\text{d}(\mathbf x, \mupar)) [\kappa] = 0 & \forall \kappa \in \mathbb U,\\
D_z\mathscr L(w(\mathbf x, \mupar); y_\text{d}(\mathbf x, \mupar)) [\zeta] = 0 & \forall \zeta \in \mathbb Y,\\
\end{cases}
\end{equation}
where $D_{y} L(w(\mathbf x, \mupar); y_\text{d}(\mathbf x, \mupar))$ is the Fr\'echet differential of the Lagrangian functional with respect to\ the state variable. The same notation is used for the other variables.
The optimality system \eqref{eq:KKT} reads: given $\mupar \in \mathcal D$, find the optimal solution $(y(\mathbf x, \mupar), u(\mathbf x, \mupar), z(\mathbf x, \mupar)) \in \mathbb Y \times \mathbb U \times \mathbb Y$ such that
\begin{equation}
\begin{cases}
 y(\mathbf x, \mupar) + D_y \mathcal E (y(\mathbf x, \mupar), u(\mathbf x, \mupar))^* (z(\mathbf x, \mupar)) = y_{\text{d}}(\mathbf x, \mupar), \\ 
 \alpha(\mupar) u(\mathbf x, \mupar) - C^* (z(\mathbf x, \mupar)) = 0, \\
 G (y(\mathbf x, \mupar)) - C(u(\mathbf x, \mupar)) = f(\mathbf x, \mupar). \\
\end{cases}
\end{equation}

It is now clear that we are dealing with a system of the form introduced in Section \ref{sec:PINN_multiple_eq} and can be written in the following compact form: given $\mupar \in \mathcal D$, find $w(\mathbf x, \mupar) \in \mathbb Y \times \mathbb U \times \mathbb Y$ such that
\begin{equation}
\label{ocp}
\mathcal G(w(\mathbf x, \mupar)) = {\mathcal F(\mathbf x, \mupar)},
\end{equation}
with
\begin{equation*}
\mathcal G(w(\mathbf x, \mupar)) \doteq \begin{bmatrix} y + D_y \mathcal E (y(\mathbf x, \mupar), u(\mathbf x, \mupar))^* (z(\mathbf x, \mupar)) \\ \alpha(\mupar) u(\mathbf x, \mupar) - C^* (z(\mathbf x, \mupar)) \\ G(y(\mathbf x, \mupar)) - C(u(\mathbf x, \mupar)) \end{bmatrix} \quad
\text{and} \quad \mathcal F(\mathbf x, \mupar) \doteq \begin{bmatrix} y_{\text{d}(\mathbf x, \mupar)} \\ 0 \\ f(\mathbf x, \mupar) \end{bmatrix}.
\end{equation*}
We want to remark that this formulation can be simplified in a non-parametrized setting, where the input of the PINN will only consist in the variable $\mathbf x$, see e.g, Remark \ref{remark:non_mu}. 
In the next Section we will present some numerical results on the application of a PI-Arch to an optimal control test problem in both parametrized and non-parametrized version.

\subsection{Numerical Results: Poisson problem}
\label{sec:numPoisson}
\begin{figure}[bt]
\begin{center}
\begin{tikzpicture}[shorten >=1pt,->,draw=black!50, node distance=\layersep]
    \tikzstyle{every pin edge}=[<-,shorten <=1pt]
    \tikzstyle{neuron}=[circle,fill=black!25,minimum size=10pt,inner sep=0pt]
    \tikzstyle{input neuron}=[neuron, fill=cyan!30];
    \tikzstyle{output neuron}=[neuron, fill=magenta!40];
    \tikzstyle{hidden neuron}=[neuron, fill=teal!50];
    \tikzstyle{annot} = [text width=4em, text centered]

    \node[input neuron, pin=left:{\footnotesize $x_0$}] (I1-1) at (0,-1 cm) {};
    \node[input neuron, pin=left:{\footnotesize $x_1$}] (I1-2) at (0,-1.5 cm) {};
    \node[input neuron, pin=left:{\footnotesize $\mu_1$}] (I1-3) at (0,-2 cm) {};
    \node[input neuron, pin=left:{\footnotesize $\mu_2$}] (I1-4) at (0,-2.5 cm) {};
    \node[input neuron, pin=left:{\footnotesize $k_1(x_0, x_1)$}] (I1-5) at (0,-3 cm) {};

    \foreach \name / \y in {1/1,2/1.5,3/2,4/2.5,5/3,6/4}
        \path[yshift=0.5cm]
            node[hidden neuron] (H1-\name) at (\layersep,-\y cm) {};

    \node (H1-7) at (\layersep, -2.9cm) {$\vdots$};

    \node[output neuron,pin={[pin edge={->}, align=left]right:{\it \footnotesize {$u(\mathbf x, \mupar)$}}}, right of=H1-3] (O1-1) {};
    \node[output neuron,pin={[pin edge={->}, align=left]right:{\it \footnotesize {$y(\mathbf x, \mupar)$}}}, right of=H1-5] (O1-2) {};


    \foreach \source in {1,...,5}
        \foreach \dest in {1,...,6}
            \path (I1-\source) edge (H1-\dest);

    \foreach \source in {1,...,6}
        \path (H1-\source) edge (O1-1);
    \foreach \source in {1,...,6}
        \path (H1-\source) edge (O1-2);
        
    \node[input neuron, right=2cm of O1-1, pin=left:{}] (I-1) {};
    \node[hidden neuron, fill=gray!10, right=0.8cm of I-1] (H-1) {$\times$};
    \node[input neuron, above=0.7cm of I-1, pin=left:{\footnotesize $\mu_2$}] (I-2) {};

    \path[draw, ->] (I-2) -| (H-1);
    \path[draw, ->] (I-1) -- (H-1);
    
    \node[output neuron,pin={[pin edge={->}, align=left]right:{\it \footnotesize {$z(\mathbf x, \mupar)$}}}, right of=H-1] (O-1) {};
    
    \path[draw, ->] (H-1) -- (O-1);
    \node[annot,above of=H1-1, node distance=1cm] (hlk) {$1^{st}$ NN};

\end{tikzpicture}
\end{center}
\caption{PI-Arch used to solve problem \eqref{eq:ocp_Poisson_PINN}.}\label{fig:pi_arch}
\end{figure}
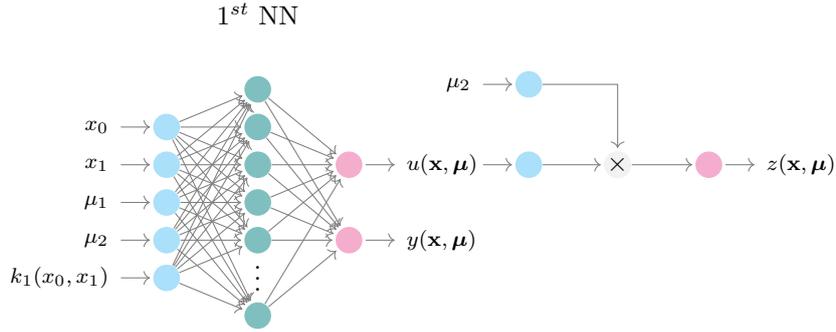

\begin{figure}
\centering
\includegraphics[width=\textwidth]{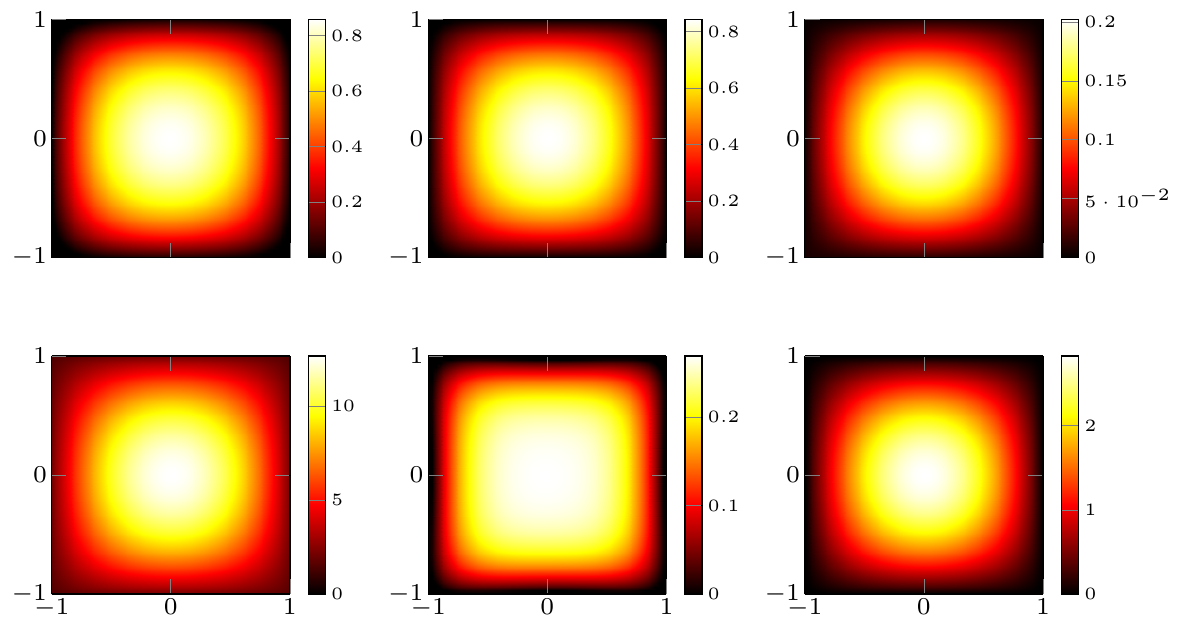}
\caption{Parametric Optimal Control Problem. \emph{\underline{Top row}}. PINN approximation for $\mupar=(3, 1)$. 
\emph{\underline{Bottom row}}. PINN approximation for $\mupar=(3, 0.01)$. \emph{Left.}  The control variable $u$. \emph{Center.} The adjoint variable $z$. \emph{Right.} The state variable $y$.}\label{fig:ocp_result}
\end{figure}

For this numerical test, we consider a steady OCP($\mupar$) in the physical domain $\Omega = [-1,1] \times [-1, 1]$. The parametric minimization problem reads: given $\mupar \doteq (\mu_1, \mu_2) \in [0.5, 3] \times [0.01, 1]$ find $(y(\mathbf x, \mupar), u(\mathbf x, \mupar)) \in H^1_0(\Omega) \times L^2(\Omega)$ such that
\begin{align}
\label{eq:ocp_Poisson_PINN}
 \nonumber
    & \min_{(y(\mathbf x, \mupar), u(\mathbf x, \mupar))} \frac{1}{2}\norm{y(\mathbf x, \mupar) - \mu_1}_{L^2(\Omega)}^2 + \frac{\mu_2}{2}\norm{u(\mathbf x, \mupar)}_{L^2(\Omega)}^2, \\
    & \text{\centering constrained to } \\ \nonumber
    & \begin{cases}
     - \Delta y(\mathbf x, \mupar) = u(\mathbf x, \mupar) & \text{ in } \Omega, \\
     y(\mathbf x, \mupar) = 0 & \text{ on } \partial \Omega.
    \end{cases}
\end{align}
Namely, we choose  $y_{\text{d}}(\mathbf x, \mupar) \equiv \mu_1$ all over the physical domain. Moreover, we also exploited a parametric penalization term.  Thus, the minimization problem seeks a distributed control term $u(\mathbf x, \mupar)$ that is able to steer the solution towards a parametric instance $\mu_1$. Applying the Lagrangian arguments of Section \ref{sec:ocp} to problem \ref{eq:ocp_Poisson_PINN}, we end up with the following optimality system once defined the adjoint variable $z(\mathbf x, \mupar) \in H^1_0(\Omega)$

\begin{equation}
\begin{cases}
y(\mathbf x, \mupar) - \Delta z(\mathbf x, \mupar) = \mu_1 & \text{ in } \Omega, \\
z(\mathbf x, \mupar) = 0 & \text{ on } \partial \Omega, \\
\mu_2 u(\mathbf x, \mupar) = z(\mathbf x, \mupar) & \text{ in } \Omega, \\
 - \Delta y(\mathbf x, \mupar) = u(\mathbf x, \mupar) & \text{ in } \Omega, \\
     y(\mathbf x, \mupar) = 0 & \text{ on } \partial \Omega.
\end{cases}
\end{equation}
Thus, we want to tackle this problem with the employment of a PI-Arch, say once again $\hat{w}$, which takes as input $(\mathbf x, \mupar)$ and gives as a result output the predicted values of the three involved variables $(y(\mathbf x, \mupar), u(\mathbf x, \mupar), z(\mathbf x, \mupar))$. Our first attempt was to use the formulation proposed in Section \ref{sec:PINN_multiple_eq}. Namely, a standard PINN was built. The prediction was based on a 3--Layers NN. The first two layers are made by $40$ neurons, while the last hidden layer had only $20$ neurons. We employed an ADAM optimizer with a Softplus activation function and LR$=0.002$. The number of the collocation points are given by: $N_p = 900$, $N_b = 200$ and $N_{\mupar} = 50$. In order to accelerate the training phase and to help the function in reaching the right boundary conditions, we employed the following extra feature:
\begin{equation}
    k_1(x_0, x_1) \doteq (1-x_0^2)(1-x_1^2).
\end{equation}
The plots in Figure \ref{fig:ocp_result} depict the results after a $10000$ epochs training. On the top row, from left to right, we observe the predicted control, adjoint and state variables for $\mupar = (3, 1)$. The results in this case were quite as expected but, once evaluated the parameter $\mupar = (3, 0.01)$ an issue came out, as represented in the bottom row of Figure \ref{fig:ocp_result}. Indeed, the standard PINN approach in this case was not able to recover the relation 
$
\mu_2 u(\mathbf x, \mupar) = z(\mathbf x, \mupar).
$
Namely, the qualitative and the quantitative behaviour of the adjoint variable $z(\mathbf x, \mupar)$ is completely different with respect to the the control variable  $u(\mathbf x, \mupar)$. For this reason, we decided to exploit the optimality equation in order to build a PI-Arch aware of such a relation between the two variables. A schematic representation of the used structure is presented in Figure \ref{fig:pi_arch}. 

\begin{figure}
\centering
\includegraphics[width=.95\textwidth]{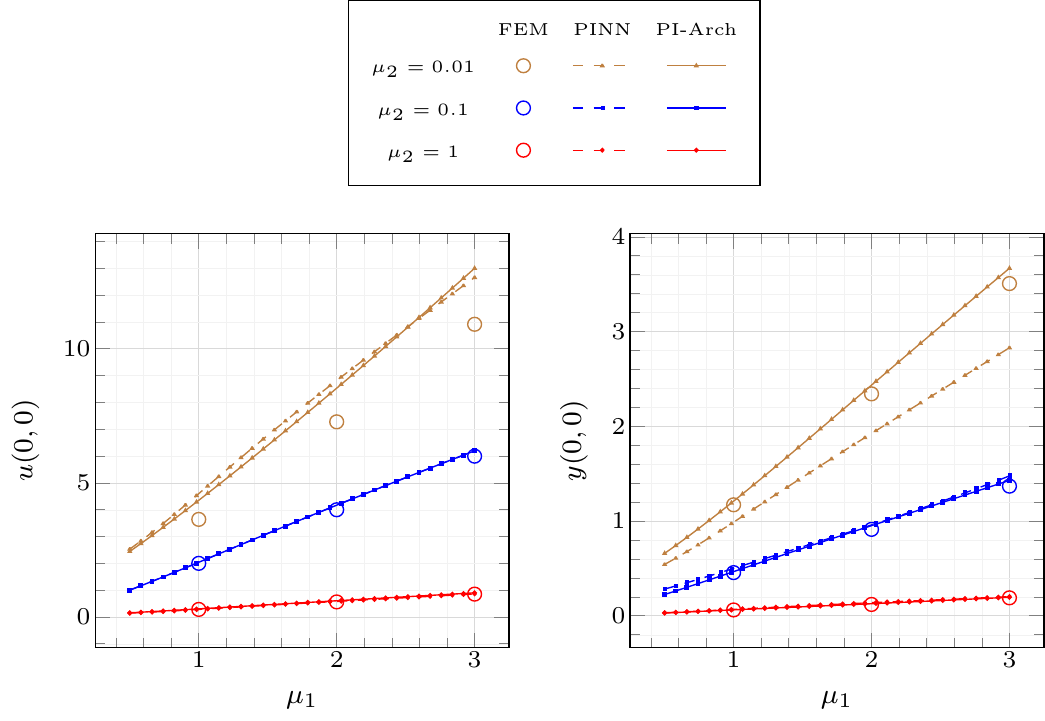}
\caption{Prediction for $\mu_2 = 1, 0.1, 0.01$ with respect to $\mu_1$ in $(x_0, x_1) = (0, 0)$ compared to the FE approximation for $\mu_1 = 1, 2, 3$. \emph{Left.} Control variable. \emph{Right.} State variable.}\label{fig:ocp_FE}
\end{figure}

Namely, a first NN approximates the 1-out given by the control and the state variables once provided the input $(\mathbf x, \mupar, k_1(x_0, x_1))$. Then, the control variable combined with the parameter $\mu_2$ is used to define the adjoint variable in another NN as $\Xi(\text{1-out}) = \mu_2 u(\mathbf x, \mupar)$, to force a meaningful physical behaviour. Figure \ref{fig:ocp_FE} shows how the PI-Arch can be effective if compared to standard PINN strategies. Namely, imposing the physics directly in the structure of $\hat w$ can be really helpful in increasing the accuracy of the prediction. Indeed, we are able not only to recover a good approximation for the case $\mupar = (3, 1)$, but also for a smaller value of the penalization parameter, i.e.\ $\mupar = (3, 0.01)$: the control and the adjoint variable are now fully coherent with the physics of the problem. To test the accuracy of the PI-Arch structure decided to compare the prediction of $\hat w$ in the domain point (0, 0) to the FE solution in $\mu_1 = 1, 2, 3$ and $\mu_2 = 1, 0.1, 0.01$. The results are depicted in Figure \ref{fig:ocp_FE}.  Comparing the standard PINN model with the PI-Arch one, we highlight that the state variable prediction --- for $\mu_2 = 0.01$ --- is closer to the FE solution with the improved architecture. 
While, we observe a good prediction for the state variable for all the values of $\mu_2$ (right plot), the quality of the approximation slightly worsen for the control variable when dealing with $\mu_2 = 0.01$.
However, the results are quite satisfactory and promising.  We remark that the PI-Arch is problem-dependent and we believe that small modifications of such a structure together with a longer training phase may lead to more reliable results. However, the topic of the right tuning of the $\hat w$ parameters and structure go beyond the goals of this work, where we want to provide a non--intrusive tool capable to answer to parametric studies needs in a small amount of time without paying in accuracy.

\begin{figure}
\centering
\includegraphics[width=\textwidth]{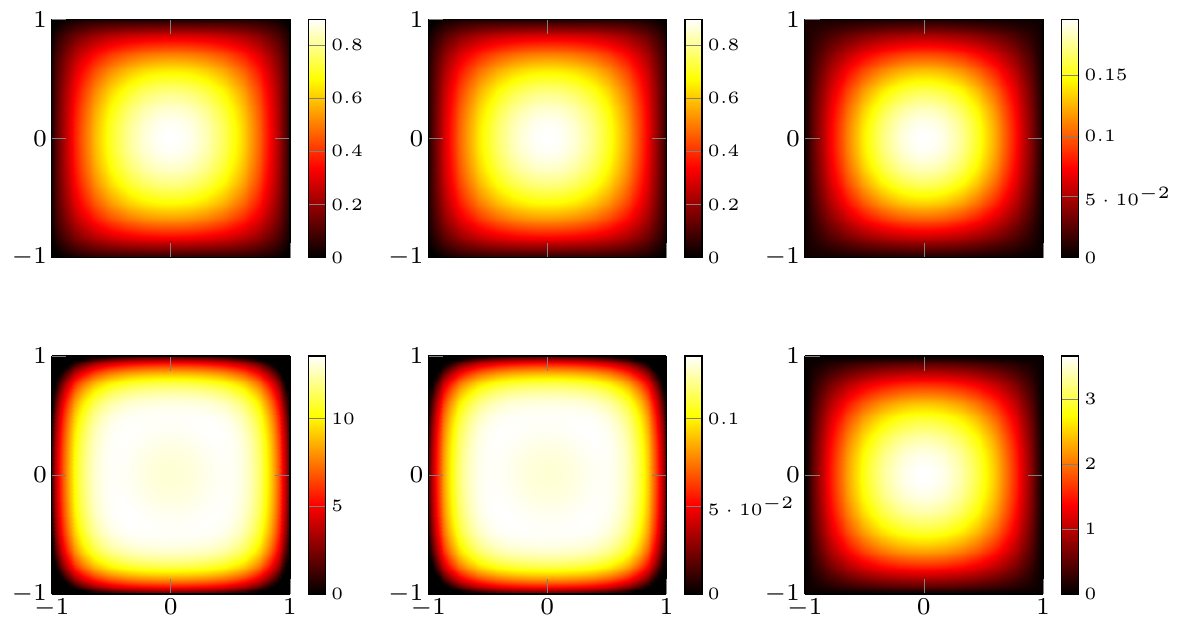}
\caption{Parametric Poisson Optimal Control Problem. \emph{\underline{Top row}}. PI-Arch approximation for $\mupar=(3, 1)$. 
\emph{\underline{Bottom row}}. PI-Arch approximation for $\mupar=(3, 0.01)$. \emph{Left.}  The control variable $u$. \emph{Center.} The adjoint variable $z$. \emph{Right.} The state variable $y$.}\label{fig:ocp_cascade_result}
\end{figure}

\subsection{Numerical Results: Stokes problem}
\begin{figure}
    \centering
    \includegraphics[width=.4\textwidth]{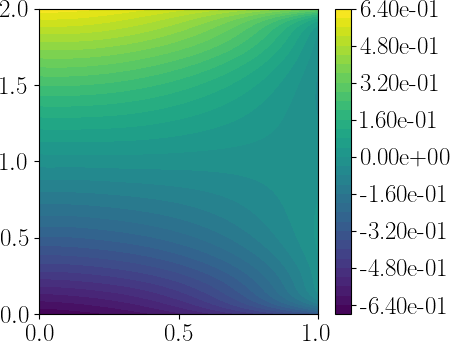}\hspace{0.8cm}
    \includegraphics[width=.4\textwidth]{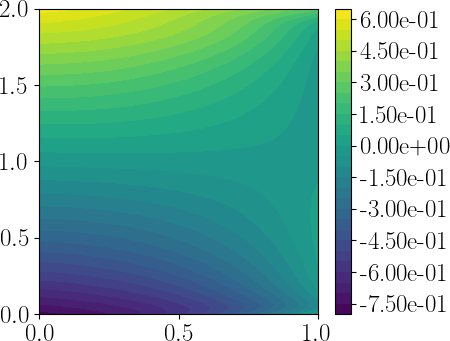}\\
    \includegraphics[width=.4\textwidth]{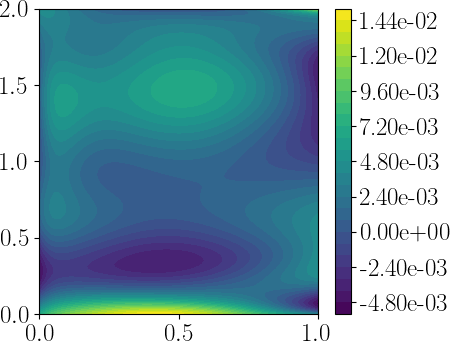}\hspace{0.8cm}
    \includegraphics[width=.4\textwidth]{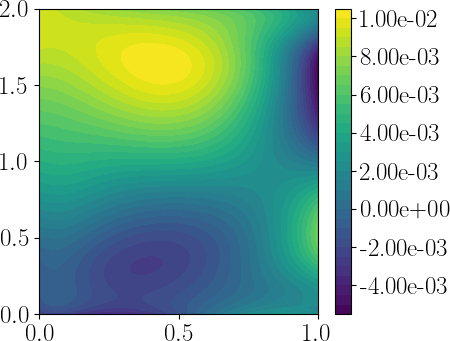}\\
    \includegraphics[width=.4\textwidth]{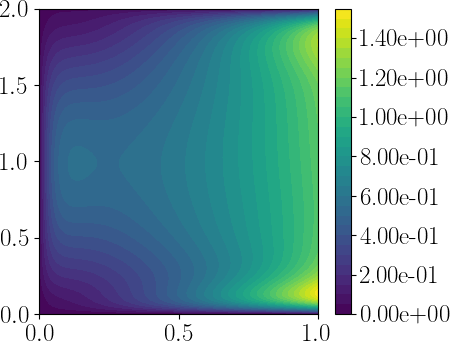}\hspace{0.8cm}
    \includegraphics[width=.4\textwidth]{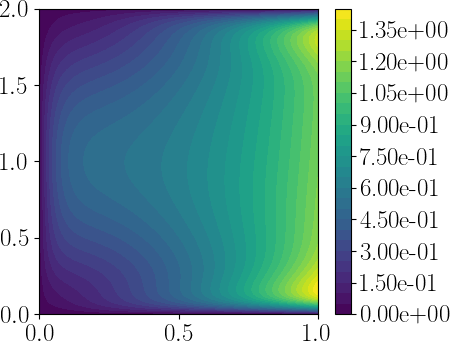}\\
    \includegraphics[width=.4\textwidth]{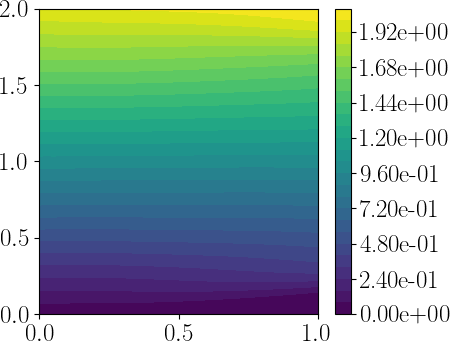}\hspace{0.8cm}
    \includegraphics[width=.4\textwidth]{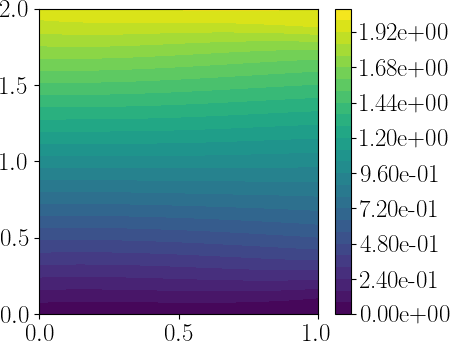}\\
    \includegraphics[width=.4\textwidth]{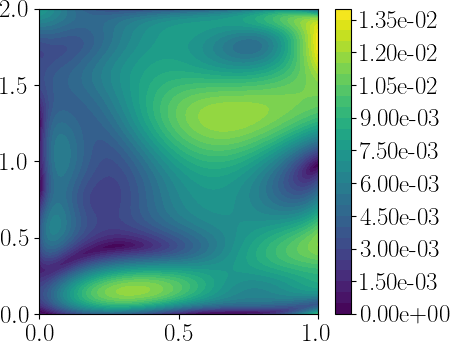}\hspace{0.8cm}
    \includegraphics[width=.4\textwidth]{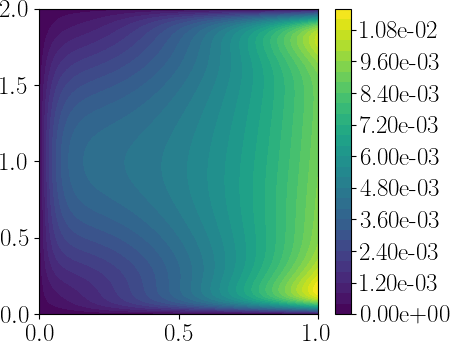}\\
    \caption{Parametric Stokes Optimal Control Problem. The field $p$, $r$, $u$, $v$ and $z$ are shown for $\mupar = 1$, from the top to the bottom.
    \emph{\underline{Left column}}. Standard FNN approximation. 
    \emph{\underline{Right column}}. PI-Arch approximation.}
    \label{fig:stokes}
\end{figure}

In the last experiment, we deal with a steady OCP($\mupar$) governed by Stokes equations. The physical domain is $\Omega = [0,1] \times [0, 2]$. We call $\Gamma_{\text{D}} \doteq \{0\}\times[0,2]$ and $\Gamma_{\text{D}} \doteq \{ 1 \} \times[0,2]$. Here, we consider a single parameter related to the forcing term: $\mupar\in [0.5, 1.5]$. The penalization parameter is fixed and we set $\alpha=0.008$ Indeed, we want to find $(v(\mathbf x, \mupar), p(\mathbf x, \mupar), u(\mathbf x, \mupar)) \in \color{blue}{[H^1_0(\Omega]^2} \times L^2(\Omega) \times [L^2(\Omega)]^2$ such that
\begin{align}
\label{eq:ocp_Stokes_PINN}
 \nonumber
    & \min_{(v(\mathbf x, \mupar), u(\mathbf x, \mupar))} \frac{1}{2}\norm{v(\mathbf x, \mupar) - x_2}_{L^2(\Omega)}^2 + \frac{\alpha}{2}\norm{u(\mathbf x, \mupar)}_{L^2(\Omega)}^2, \\
    & \text{\centering constrained to } \\ \nonumber
    & \begin{cases}
     - 0.1\Delta v(\mathbf x, \mupar) + \nabla p(\mathbf x, \mupar) = f(\mathbf x, \mu_1) + u(\mathbf x, \mupar) & \text{ in } \Omega, \\
     \nabla \cdot v(\mathbf x, \mupar) = 0 & \text{ in } \Omega, \\
     v(\mathbf x, \mupar)_1 = x_2 \text{ and } v(\mathbf x, \mupar)_2 = 0 & \text{ on } \Gamma_{\text{D}}, \\
     \displaystyle -p(\mathbf x, \mupar)\mathbf{n}_1 + 0.1 \frac{\partial v(\mathbf x, \mupar)_1}{\partial \mathbf n_1} \text{ and } v(\mathbf x, \mupar)_2 =0   & \text{ on } \Gamma_{\text{N}}. 
    \end{cases}
\end{align}
The subscripts $_1$ and $_2$, indicate the first and the second component of a vector field, respectively. The formulation is totally similar to the one proposed in Section \ref{sec:numPoisson} we can apply standard Lagrangian arguments and the minimization problem translates in the solution of the following system:

\begin{equation}
\begin{cases}
\label{eq:ocp_Stokes_aadj}
     - 0.1\Delta z(\mathbf x, \mupar) + \nabla r(\mathbf x, \mupar) = \begin{bmatrix} x_2 - v(\mathbf x, \mupar) _1 \\ 0
     \end{bmatrix} & \text{ in } \Omega, \\
     z(\mathbf x, \mupar)_1 = z(\mathbf x, \mupar)_2 = 0 & \text{ on } \Gamma_{\text{D}}, \\
      \nabla \cdot z(\mathbf x, \mupar) = 0 & \text{ in } \Omega, \\
      \displaystyle -r(\mathbf x, \mupar)\mathbf{n}_1 + 0.1 \frac{\partial z(\mathbf x, \mupar)_1}{\partial \mathbf n_1} \text{ and } z(\mathbf x, \mupar)_2 =0   & \text{ on } \Gamma_{\text{N}}. \\
      \alpha u(\mathbf x, \mupar) = z(\mathbf x, \mupar)  & \text{ in } \Omega, \\
     - 0.1\Delta v(\mathbf x, \mupar) + \nabla p(\mathbf x, \mupar) = f(\mathbf x, \mu_1) + u(\mathbf x, \mupar) & \text{ in } \Omega, \\
     \nabla \cdot v(\mathbf x, \mupar) = 0 & \text{ in } \Omega, \\
     v(\mathbf x, \mupar)_1 = x_2 \text{ and } v(\mathbf x, \mupar)_2 = 0 & \text{ on } \Gamma_{\text{D}}, \\
     \displaystyle -p(\mathbf x, \mupar)\mathbf{n}_1 + 0.1 \frac{\partial v(\mathbf x, \mupar)_1}{\partial \mathbf n_1} \text{ and } v(\mathbf x, \mupar)_2 =0   & \text{ on } \Gamma_{\text{N}},
\end{cases}
\end{equation}
where $(z(\mathbf x, \mupar), r(\mathbf x, \mupar)) \in \times [H^1_0(\Omega]^2 \times L^2(\Omega)$ is the adjoint variable for this problem. Also in this case, using the PI-Arch structure $\hat{w}$ led to advantages in terms of accuracy in the prediction of the five variables, given $(\mathbf x, \mupar)$ as inputs. The results with respect to the standard PINN proposed in Section \ref{sec:PINN_multiple_eq} are depicted in Figure~\ref{fig:stokes}. We build a standard PINN with 4--Layers NN where all the hidden layers are made of $40$ neurons. For the prediction, we exploited an ADAM optimizer with a Softplus activation function and LR$=0.003$. The collocation points are collected using a latin hypercube strategy, for a total of $N_p = 400$, $N_b = 1800$ and $N_{\mupar} = 10$.\\
No extra-features have been applied in this case, since $x_2$, i.e.\ the Dirichlet boundary condition for the first component of the state velocity, is an input itself, already.
The plots in Figure  depict the results after a $10000$ epochs training. On the top row, from left to right, we observe the predicted control, adjoint and state variables for $\mupar = 1$. In this case, an issue came out, as represented in the bottom row of Figure \ref{fig:ocp_result}. Indeed, the standard PINN approach was not able to recover the relation 
$
\alpha u(\mathbf x, \mupar) = z(\mathbf x, \mupar).
$
Namely, the qualitative and the quantitative behavior of the adjoint variable $z(\mathbf x, \mupar)$ is completely different with respect to the control variable  $u(\mathbf x, \mupar)$. 

To avoid this issue, we explicit the optimality equation in the PI-Arch structure, as already done in the previous test case. The strategy is totally analogous: a first NN (with the same parameter of the standard PINN described above) predicts the 1-out that predicts the control, the state variables, and the adjoint variable $r(\mathbf x, \mupar)$ from the input $(\mathbf x, \mupar)$. Then, the adjoint variable $z(\mathbf x, \mupar)$ is the result of another NN defined as $\Xi(\text{1-out}) = \alpha u(\mathbf x, \mupar)$: we are imposing the physical constraint directly. 
Figure~\ref{fig:stokes} shows how the PI-Arch can be effective if compared to standard PINN strategies. Namely, imposing the physics directly in the structure of $\hat w$ can be really helpful in increasing the accuracy of the prediction. Indeed, we are able to recover a good approximation for the case $\mupar = 1$, even using the penalization factor $\alpha = 0.008$, which implies different order of magnitude in the predicted unknowns. The control and the adjoint variable are now fully coherent with the physics of the problem, contrary to the standard architecture.
The results are quite satisfactory and promising, even if it must be said that more consolidated frameworks --- e.g. finite elements --- are surely able to achieve better accuracy. We remark that the PI-Arch is problem-dependent and we believe that small modifications of such a structure together with a longer training phase may lead to more reliable results. However, the topic of the right tuning of the $\hat w$ parameters and structure go beyond the goals of this work, where we want to provide a non--intrusive tool capable to answer to parametric studies needs in a small amount of time without paying in accuracy.

\section{Conclusions}
\label{sec:conclusions}
We propose a PINN extension to parametric settings in a real-time and many-query context. The main goal of this contribution was to adapt the PINN paradigm to the need for fast simulations for several parametric instances in a small amount of time. Indeed, PINN is a valuable tool in this framework. After the training phase, several $\mupar$-dependent outputs can be computed in a rapid sequence. However, another aspect to consider is how long this training phase could be. To overcome a possibly unbearable PINN training, we used the extra features: namely, an augmented input drives the NN towards the loss minimum in a smaller number of epochs. Furthermore, we propose the PI-Arch technique to build tailored NNs to deal with complex problems. The methodologies have been tested both on forward problems and on more complex systems of equations, such as OCP($\mupar$)s.  
Besides the promising results we have shown, many open questions and points remain of interest. \\
First of all, we would like to treat more complex problems in the OCP($\mupar$)s framework. Indeed, optimal control is a mathematical tool suited in many data-based contexts, such as inverse problems and data assimilation, to bridge the gap between data and mathematical modelling. Thus, the combination with PINNs seems a natural choice to integrate some data information in the problem at hand without spoiling its physical meaning. The presented results are just a first step that moves in the direction of cases of actual interest. \\
Another improvement concerns the choice of the extra features and the PI-Arch structure. In this work, they are assumed to be known. We are aware that this is not the case for many real applications. Thus, many natural questions might arise. Among them: is there a way to appropriately choose the features? Is there a way to learn them? What is the best way to connect the PINN architecture with this knowledge?\\
For sure, the presented methodology enhanced with the proposed improvements would pave the way to the use of PINN in very complex settings, ranging from several application fields in science and industry.

\section*{Acknowledgements}
This work was partially supported by European Union Funding for Research and Innovation ---
Horizon 2020 Program --- in the framework of European Research Council
Executive Agency: H2020 ERC CoG 2015 AROMA-CFD project 681447 ``Advanced
Reduced Order Methods with Applications in Computational Fluid Dynamics'' P.I.
Gianluigi Rozza.
The work was also supported by INdAM-GNCS: Istituto Nazionale di Alta Matematica --– Gruppo Nazionale di Calcolo Scientifico. We thank the support from PRIN project NA-FROM-PDEs (MIUR).

\appendix
{\section{\A{Some additional results}}
\label{sec:results_laplace_burgers}
We are going to discuss the advantages of using extra features to accelerate the training phase reaching the best approximation results for the physical phenomenon one is dealing with, \A{not necessarily in the constrained optimization framework}. The paradigm of PINN is to add the physical information you have on the system to reach solutions that are related to PDEs models. The employment of extra features totally complies with the PINN fundamentals: add previous knowledge to your system to reach a more reliable solution and, most of all, in a smaller training time. Finally, we will test the PINN methodology in a parametrized setting, too.
\subsection{Poisson problem}
The first problem we are going to tackle is a Poisson problem. The test case is a slightly modified version of the problem presented in~\cite{Atangana2013}. We consider the two-dimensional steady case, given by the following continuous model:
\begin{equation}
\label{eq:Poisson}
    \begin{cases}
     \Delta w(\mathbf x) = f(\mathbf x) & \text{ in } \Omega, \\
    w(\mathbf x) = 0 & \text{ on } \partial \Omega,
    \end{cases}
\end{equation}
where $\Omega$ is the unit square $[0,1]\times[0,1]$, a point in $\Omega$ is $\mathbf x = (x_0, x_1)$  and the forcing term is defined as 
$$
f(\mathbf x) \doteq 
\sin(\pi x_0)\sin(\pi x_1).
$$  
The Poisson problem \eqref{eq:Poisson} has the following analytical solution:
\begin{equation}
    w(\mathbf x) = -\frac{\sin(\pi x_0)\sin(\pi x_1)}{2\pi^2}.
\end{equation}
After having defined the residual as we did in \eqref{eq:res}, we choose $N_p = 100$ and $N_b = 40$ following an equispaced rule. We deal with the problem through a 2--Layer NN of $10$ neurons for each layer. As optimizer, we exploited ADAM~\cite{adam} with a learning rate (LR) fixed at $0.003$ and a Softplus activation function. We show three numerical examples to state how the features can be useful in order to accelerate the training procedure and reach faster convergence rates.
\begin{itemize}
    \item[$\circ$](\emph{No extra features}) First of all, Figure \ref{fig:poisson_no_feat_visual} shows a comparison between the analytical solution of problem \eqref{eq:Poisson} and the PINN solution. They coincide. The accuracy of the PINN is assured also by the right plot representing the absolute value of the pointwise difference between the solutions, say the \emph{pointwise error}, that reaches the value of $10^{-3}$ after $1000$ epochs.  
    \item[$\circ$](\emph{With extra features}) For the same NN architecture, we tested the effect of the extra features in this context. Guided by the knowledge on the forcing term, we enforced the input with the extra feature given by
    \begin{equation}
    \label{eq:k_1}
        k_1(x_0, x_1) \doteq \sin(x_0 \pi)\sin(x_1 \pi).
    \end{equation}
    From Figure \ref{fig:poisson_feat_visual}, it is clear that  we are gaining an order of magnitude in the relative error with respect to standard PINN approach (we refer to the right plot).
    \item[$\circ$](\emph{Learnable extra features}) This phenomenon is highlighted in the third example, where we exploit a \emph{learnable feature}. Namely, we enforce the following information on the system:
    \begin{equation}
    \label{eq:k_1_learn}
        k_1(x_0, x_1) \doteq \beta_0 \sin(\alpha_0 x_0 + \gamma_0)\beta_1\sin(\alpha_1 x_1 + \gamma_1),
    \end{equation}
    where $\alpha_0, \alpha_1, \beta_0, \beta_1, \gamma_1$ and $\gamma_2$ are real values to be determined and learned by the NN. 
\end{itemize}
\begin{figure}[H]
\includegraphics[width=\textwidth]{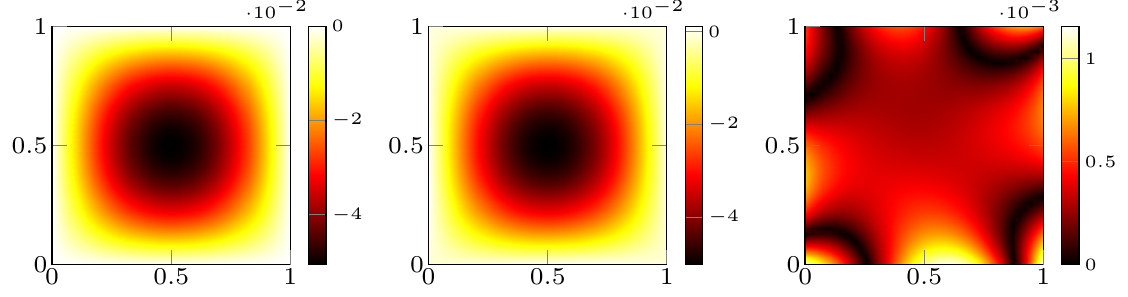}
\caption{Poisson without extra features. \emph{Left.} Analytical solution. \emph{Center.} PINN solution. \emph{Right.} Pointwise error between the analytical and the PINN solution.} \label{fig:poisson_no_feat_visual}
\end{figure}

\begin{figure}[H]
\includegraphics[width=\textwidth]{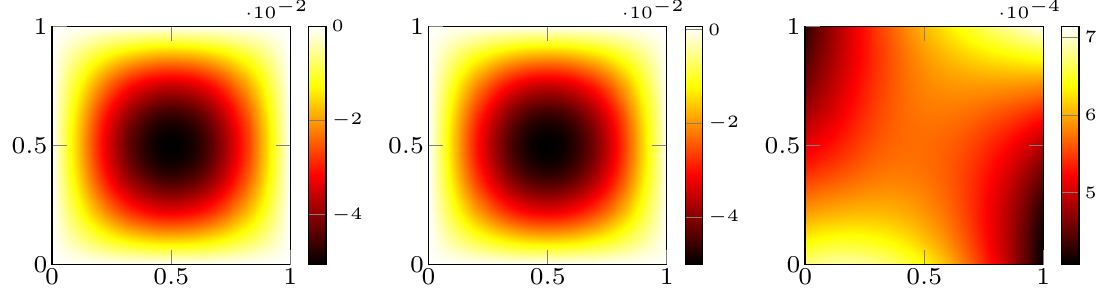}
\caption{Poisson with extra features. \emph{Left.} Analytical solution. \emph{Center.} PINN solution. \emph{Right.} Pointwise error between the analytical and the PINN solution.}\label{fig:poisson_feat_visual}
\end{figure}

\begin{figure}[H]
\includegraphics[width=\textwidth]{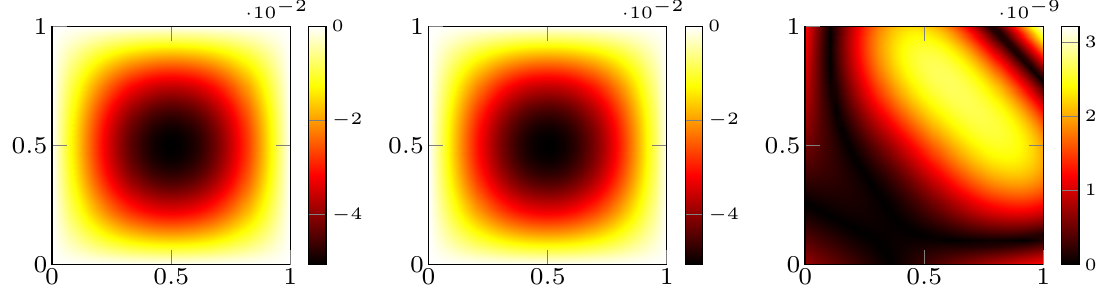}
\caption{Poisson with learnable extra features. \emph{Left.} Analytical solution. \emph{Center.} PINN solution. \emph{Right.} Pointwise error between the analytical and the PINN solution.} \label{fig:poisson_learn_feat_visual}
\end{figure}

The advantages of the latter approach is remarkable since, after $1000$ epochs, we reach an absolute pointwise error of $10^{-9}$, exploiting a fixed LR$=0.008$. In this test case, we keep the same number of boundary and internal points as before ($N_b = 40$ and $N_p = 100$) but we remove all the hidden layers in the model, converging to a simple architecture constituted by only input and output layers.
This of course has been possible since the learnable feature we provided can be fit to the analytical solution of the problem, without the needed of using nonlinear activation functions. However, the usage of such features does not affect the time of the training phase which still remain in the order of minutes.
From the three numerical examples we reported, a remarkable improvement is noticed in terms of relative error for a fixed number of epochs. This is also confirmed by Figure \ref{fig:poisson_loss}, where the loss values of the different approaches is depicted. First of all, the learnable extra feature is capable to reach machine precision loss around 1500 epochs, where the standard PINN loss is stuck around $10^{-3}$. The decay of the loss is helped even by the extra feature \ref{eq:k_1}, reaching the threshold of $10^{-3}$ for only 500 epochs. These experiments heuristically prove how important is to add all the information you have on your system in order to have a reliable NN prediction in a small amount of training time.
\begin{figure}[H]
\includegraphics[width=\textwidth]{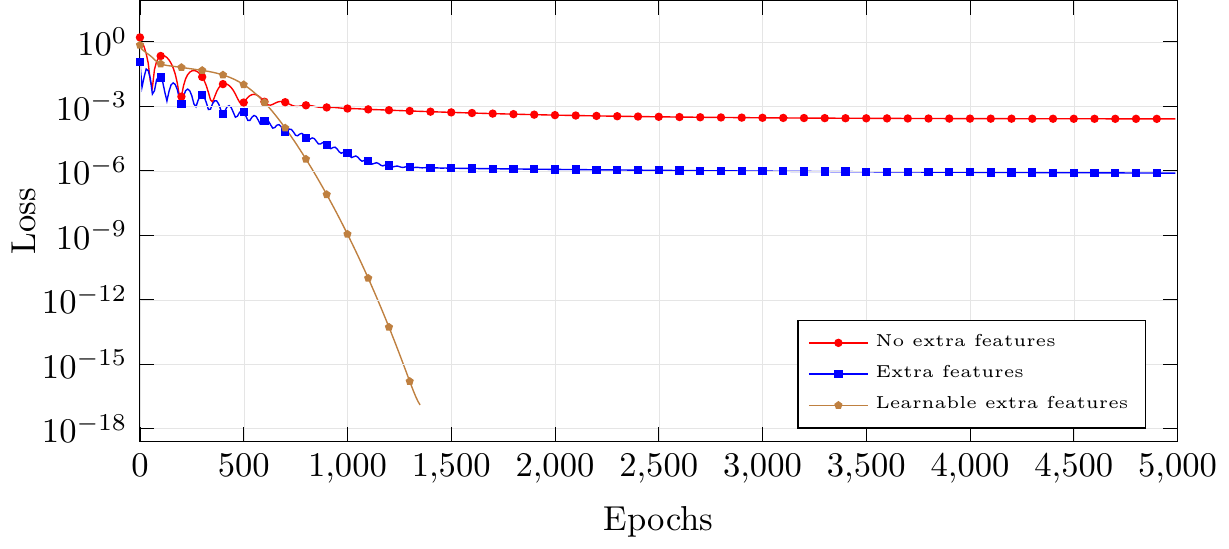}
\caption{Poissoin problem. Comparison between the three approaches (no extra features, with extra features and with learnable extra features) in terms of loss value with respect to the number of epochs.}\label{fig:poisson_loss}
\end{figure}

\subsubsection{Poisson problem with a different forcing term}
The previous results have shown a remarkable gain in the usage of extra features. It must be said that in that specific case the forcing term we impose is linearly related with the analytical solution. It implies that, using such forcing term as an extra feature, the input-output mapping we want to discover during the learning procedure results simplified.

To better test the generability of the framework, we keep the original model \eqref{eq:Poisson} varying the forcing term such as:
$$
f(\mathbf x) \doteq 
-2(x_1(1-x_1)+x_0(1-x_0)).
$$  
The analytical solution of the problem become so:
\begin{equation}
    w(\mathbf x) = x_0(1-x_0)x_1(1-x_1).
\end{equation}
Regarding the neural network architecture, we keep the same setting of the previous experiments: $2$ hidden layers composed by $10$ neurons each, with the Softplus activation function. The optimization has carried out by means of the ADAM algorithm with learning rate $\text{LR} = 0.003$. The cardinality of the point set where we evaluate the residuals are $N_p = 100$ and $N_b = 40$, distributed using a cartesian grid.
As before, we solve the problem using PINN, initially without any extra features, then using the forcing term as extra feature. Formally:
\begin{equation}
k_1(\mathbf x) \doteq 
-2(x_1(1-x_1)+x_0(1-x_0)).
\end{equation}
The comparison between the two approaches --- with and without the feature --- aims to empirically prove the effectiveness of the extra features when the latter are not linearly related with the solution of the problem. Figure~\ref{fig:poisson2_visual} presents the results obtained after a fixed amount of epochs ($\num{10000}$). The error distribution shows a better accuracy by involving the extra feature, which is demonstrated also by the visual comparison of the PINN solution with respect to the analytical solution. The error with extra features reaches an order of magnitude less than the error without features. This behaviour is also confirmed by the loss function in the two cases represented in Figure~\ref{fig:poisson2_loss}: the optimization convergence is increased by using the forcing term as an additional features also in this case.
\begin{figure}[t]
\includegraphics[width=\textwidth]{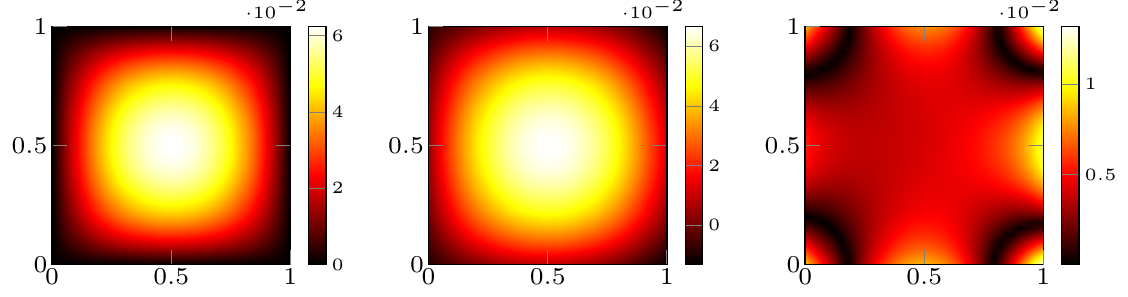}
\includegraphics[width=.99\textwidth]{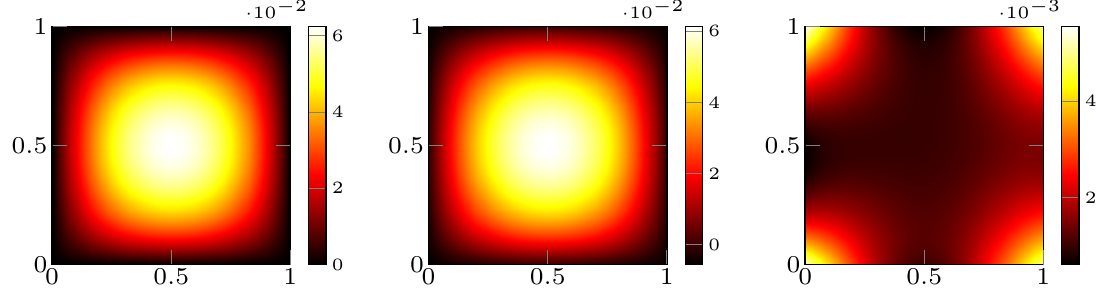}
\caption{Second Poisson problem. \emph{\underline{Top row}}. Representative solution and pointwise error without extra features. 
\emph{\underline{Bottom row}}. Representative solution and pointwise error with extra features. \emph{Left.} Analytical solution. \emph{Center.} PINN solution. \emph{Right.} Pointwise error between the analytical and the PINN solution.}\label{fig:poisson2_visual}
\end{figure}
\begin{figure}[H]
\includegraphics[width=\textwidth]{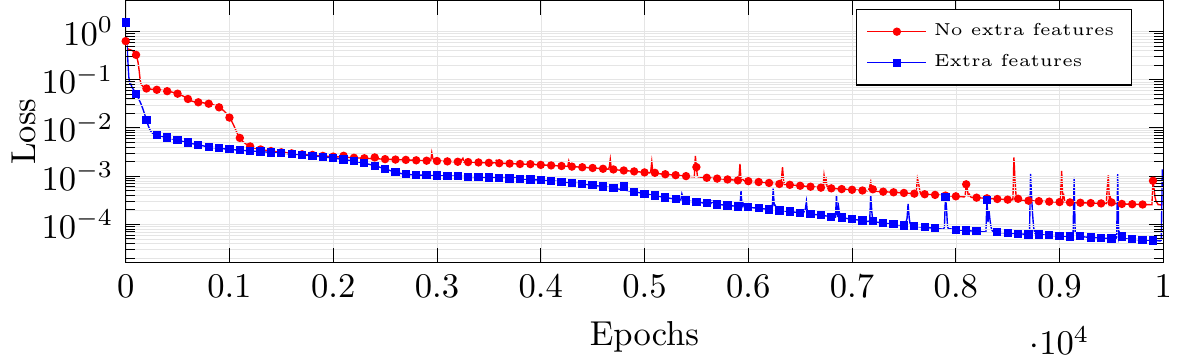}
\caption{Second Poissoin problem. Comparison between the two approaches (no extra features and with extra features) in terms of loss value with respect to the number of epochs.}\label{fig:poisson2_loss}
\end{figure}

\subsection{Burgers problem} 
\definecolor{color1}{rgb}{0.75,0,0.75}
\definecolor{color0}{rgb}{0,0.75,0.75}

We are now going to test the example presented in \cite{raissi2019physics} concerning Burgers’s equation. Also in this case, our goal is to state the advantages of employing some previous knowledge on the system enriching the input through some extra features. Considering $\mathbf x = (x, t)$, the PDE we are dealing with is:
\begin{equation}
\label{eq:burgers}
\begin{cases}
   \displaystyle  w(\mathbf x)_t + w(\mathbf x)w(\mathbf x)_x - \frac{0.01}{\pi}w(\mathbf x)_{xx} \color{blue}{ = 0} & \text{ for } x \in [-1, 1] \text{ and } t \in [0,1], \\
   w(1, t) = w(- 1, t) = 0, & \\
   w(x, 0) =  - \sin (\pi x).
    \end{cases}
\end{equation}
\begin{figure}
\includegraphics[width=\textwidth]{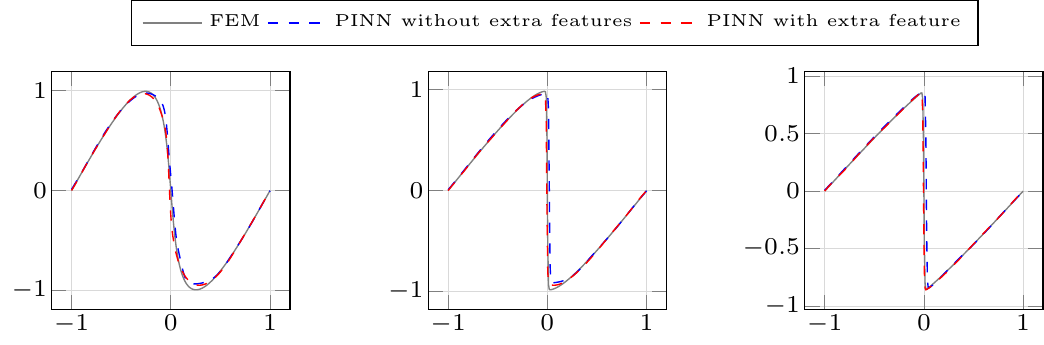}
\caption{Burgers problem with and without extra features after \num{10000} epochs. \emph{Left.} $t = 0.25$. \emph{Center.} $t=0.5$. \emph{Right.} $t = 0.75$.} \label{fig:burger_10k}
\end{figure}
\begin{figure}
\includegraphics[width=\textwidth]{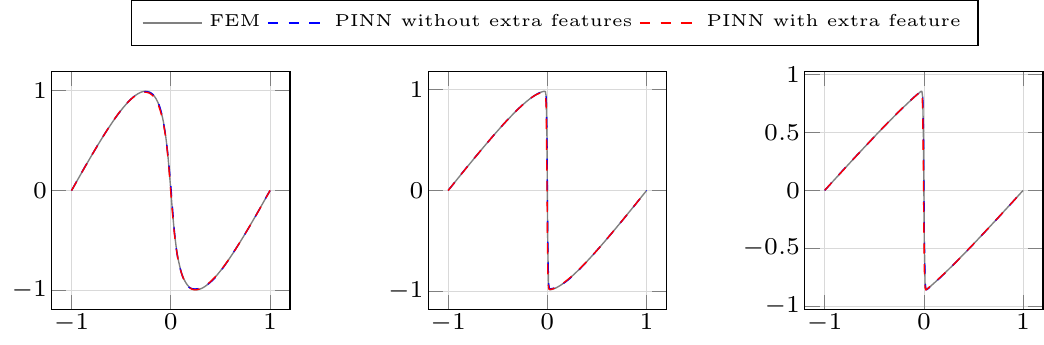}
\caption{Burgers problem with and without extra features when loss is less than \num{0.0001}. \emph{Left.} $t = 0.25$. \emph{Center.} $t=0.5$. \emph{Right.} $t = 0.75$.} \label{fig:burger_1e5}
\end{figure}

\begin{figure}
\includegraphics[width=\textwidth]{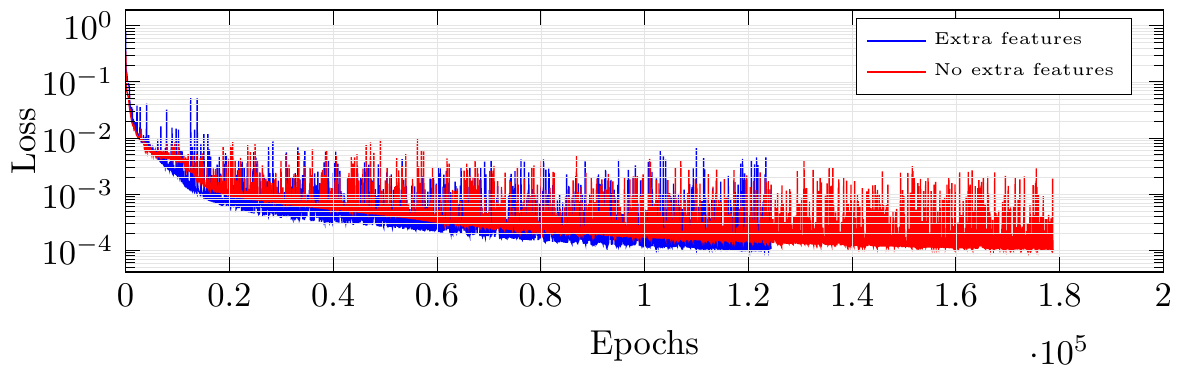}
\caption{Burgers problem. \emph{Left}. Comparison between the two approaches (no extra features and with extra features) in terms of loss value with respect to the number of epochs. \emph{Right}. Representation of the boundary and internal collocation points.}\label{fig:Burgers_loss}
\end{figure}
Let us fix the following architecture for the PINN we want to approximate the physical phenomenon at hand.  We employed a 3--Layer NN, where the number of neurons --- from first to last layer --- are equal to $20, 10$ and $5$. As the previous test cases, we used ADAM optimizer with LR$=0.006$ and a hyperbolic tangent as an activation function. The $N_p = \num{8000}$ internal points are distributed using a latin hyper-cube sampling, whereas the $N_b = 150$ boundary points are selected with a uniform random distribution. Figures~\ref{fig:burger_10k} and~\ref{fig:burger_1e5} show the comparison between standard PINN without features and with the extra features, after a fixed number of epochs (\num{10000}) and once the loss has reached the prescribed tolerance (\num{0.0001}). 

Let us comment on the results we obtained. In Figure \ref{fig:burger_10k} we show the improvements obtained by exploiting the following feature (that is nothing but the initial condition we impose in the problem) in the training procedure
    \begin{equation}
    \label{eq:k_1_burgers}
        k_1(x) \doteq \sin(x \pi).
    \end{equation}
    
It is possible to note that we gain accuracy at instants $t \in \{0.25, 0.05, 0.75\}$. The advantage of involving an extra features are however less remarkable than in the previous numerical experiments: our hypothesis is that the initial condition does not present the shock we expect after a certain time in the Burgers solution, requiring a great number of epochs to learn it accurately. Consistently, Figure~\ref{fig:Burgers_loss} presents such marginal differences also in the loss trend. The use of the extra feature allows us to reach the imposed tolerance (\num{0.0001}) after $\num{120000}$, contrarily to standard PINN formulation that requires $\num{180000})$ for the same threshold. Of course, selecting a tolerance threshold as the stopping criteria of the learning procedure makes both the approaches --- with and without extra features --- converging to the truth solution (Figure~\ref{fig:burger_1e5}).
Also in this more complicated case, we can state that introducing the feature information can be of advantage in order to reduce the training time. We would like to stress that in term of the pointwise error, the results might be improved choosing tailored features of working with more hidden layers and/or neurons. However, this went beyond the aim of this experiment.

\subsection{Parametric Poisson problem}
This Section focuses on the application of PINN in a parametric context. We propose the following problem: given a parametric instance $\mupar \doteq (\mu_1, \mu_2) \in [-1, 1] \times [-1,1]$, find the solution of 
\begin{equation}
\label{eq:Poisson_par}
    \begin{cases}
    - \Delta w(\mathbf x, \mupar) = f(\mathbf x, \mupar) & \text{ in } \Omega, \\
    w(\mathbf x, \mupar) = 0 & \text{ on } \partial \Omega,
    \end{cases}
\end{equation}
where $\Omega = [-1, 1] \times [-1, 1]$ and the forcing term is
\begin{equation}
    f(\mathbf x, \mupar) = e^{-2((x_0 - \mu_1)^{2} + (x_1 - \mu_1)^{2})}.
\end{equation}
Our aim is to recover the solution $w(\mathbf x, \mupar)$ for several parametric instances after a training phase which takes into consideration how the system changes with respect to these physical parameters. The information is given by the modified loss \eqref{eq:global_loss}. In order to achieve this prediction goal, we used an ADAM optimizer over a 3--Layers NN of $20$ neurons for each hidden layer. We chose a Softplus activation function and LR$=0.03$. The number of the collocation points are given by: $N_p = 400$, $N_b = 80$ and $N_{\mupar} = 40$, all considered with an equispaced distribution. The plots in Figure \ref{fig:poisson_par} shows satisfactory results for a $1000$ epochs training. It depicts the behaviour of two parametric solution, i.e. $w(\mathbf x, (-0.8, -0.8))$ and
$w(\mathbf x, (0.8, 0.8))$, respectively in the top and bottom row. For both the values we show a FE numerical simulation and its PINN counterpart together with the pointwise error which stays below the value of $3 \cdot 10^{-2}$.
Also in this case, the input features of the NN model are enriched by the forcing term of the equations in order to improve the training. The extra feature is defined as
$k_1(x) \doteq e^{-2((x_0 - \mu_1)^{2} + (x_1 - \mu_1)^{2})}$.
\begin{figure}
\includegraphics[width=\textwidth]{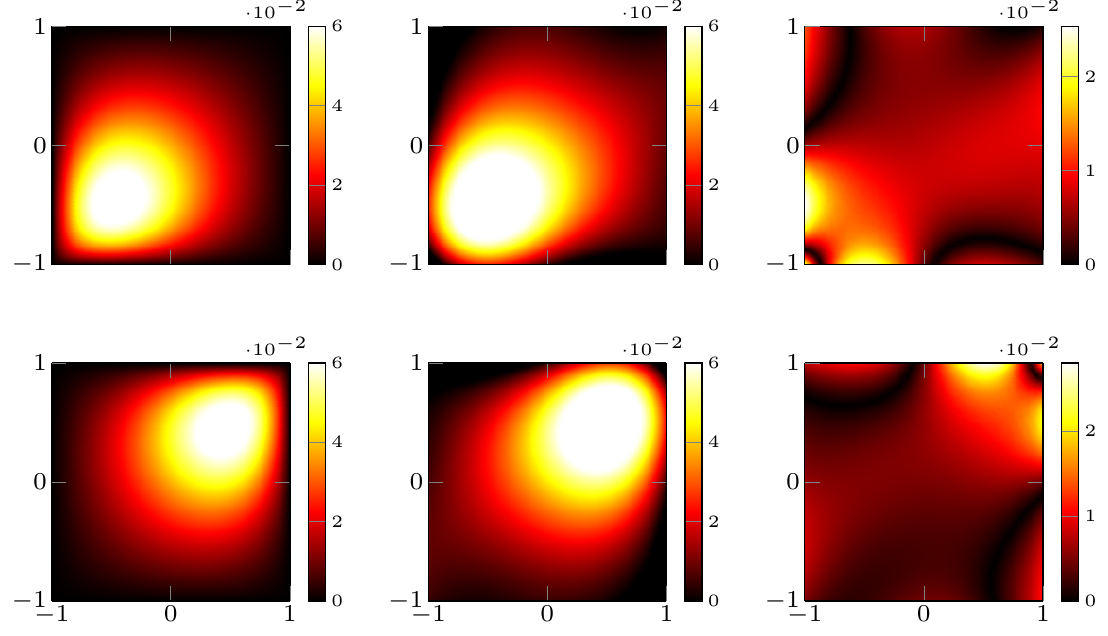}
\caption{Parametric Poisson. \emph{\underline{Top row}}. Representative solution and pointwise error for $\mupar=(-0.8, -0.8)$. 
\emph{\underline{Bottom row}}. Representative solution and pointwise error for $\mupar=(0.8, 0.8)$. \emph{Left.} Numerical solution. \emph{Center.} PINN solution. \emph{Right.} Pointwise error between the numerical and the PINN solution.} \label{fig:poisson_par}
\end{figure}

\bibliographystyle{abbrv}
\bibliography{biblio}

\end{document}